\documentclass[letterpaper,10pt,conference]{ieeeconf}

\IEEEoverridecommandlockouts
\usepackage{times}
\usepackage{amsmath,amssymb,amsfonts}

\usepackage{amsthm}
\usepackage{graphicx}
\usepackage{booktabs}
\usepackage{cite}
\usepackage[ruled]{algorithm2e}
\usepackage{listings}
\usepackage{siunitx}
\usepackage[table]{xcolor}
\usepackage{hyperref}
\usepackage{float}
\usepackage{subcaption}
\usepackage{soul}
\usepackage{multirow}
\usepackage{threeparttable}
\usepackage[most]{tcolorbox}

\theoremstyle{definition}

\theoremstyle{remark}

\theoremstyle{plain}

\sethlcolor{yellow}

\title{\LARGE A Semantic Observer Layer for Autonomous Vehicles: Pre-Deployment Feasibility Study of VLMs for Low-Latency Anomaly Detection}

\author{Kunal Runwal$^{2}$, Swaraj Gajare$^{2}$, Daniel Adejumo$^{2}$, Omkar Ankalkope$^{2}$,\\
        Siddhant Baroth$^{3}$, and Aliasghar Arab$^{1,2*}$%
\thanks{$^{1}$Aliasghar Arab is with The City College of New York, Grove School of Engineering,
        New York, NY 10031, USA.
        {\tt\small aarab@ccny.cuny.edu}}%
\thanks{$^{2}$Kunal, Swaraj, Daniel, Omkar, and Aliasghar are with the Department of Mechanical
        and Aerospace Engineering, Tandon School of Engineering, New York University,
        6 MetroTech Center, Brooklyn, NY 11201, USA (ASAS-Labs).
        {\tt\small aliasghar.arab@nyu.edu}}%
\thanks{$^{3}$Siddhant Baroth is with the Department of Electrical and Computer Engineering,
        Tandon School of Engineering, New York University,
        6 MetroTech Center, Brooklyn, NY 11201, USA (ASAS-Labs).}%
}

\begin{document}
\maketitle
\vspace{-1em}

\begin{abstract}
Semantic anomalies—context-dependent hazards that pixel-level detectors cannot reason about—pose a critical safety risk in autonomous driving. We propose a \emph{semantic observer layer}: a quantized vision-language model (VLM) running at 1--2\,Hz alongside the primary AV control loop, monitoring for semantic edge cases, and triggering fail-safe handoffs when detected. Using Nvidia Cosmos-Reason1-7B with NVFP4 quantization and FlashAttention2, we achieve $\sim$500\,ms inference---a $\sim$50$\times$ speedup over the unoptimized FP16 baseline (no quantization, standard PyTorch attention) on the same hardware---satisfying the observer timing budget. We benchmark accuracy, latency, and quantization behavior in static and video conditions, identify NF4 recall collapse (10.6\%) as a hard deployment constraint, and a hazard analysis mapping performance metrics to safety goals. The results establish a pre-deployment feasibility case for the semantic observer architecture on embodied-AI AV platforms.
\end{abstract}

\section{INTRODUCTION}
\label{sec:introduction}
\subsection{Motivation}
        Undetected semantic anomalies are a direct hazard in autonomous driving: an AV that cannot distinguish a deflated ball on the road from a shadow, or misreads traffic lights on a transport truck, may take incorrect or fatal actions~\cite{vojir2021road, tian2024latency, ZhengVideoBasedTA}. The broader robotics community has recognized the same need—recent work on spatial awareness~\cite{nwankwo2025envodat}, traversability~\cite{liang2025gnd, hanson2022vast}, and embodied navigation~\cite{duan2026causalnav} all confirm that safe operation in unstructured outdoor environments demands context-aware semantic reasoning, not just pixel-level detection. Yet none of these systems deliver this reasoning within the sub-second latency budget required for on-vehicle deployment. Building a framework that is both semantically capable \emph{and} fast enough to act in time is therefore the central challenge this work addresses.

\subsection{Problem Statement}
    \begin{figure}[ht!]
        \centering
        \includegraphics[width=\linewidth]{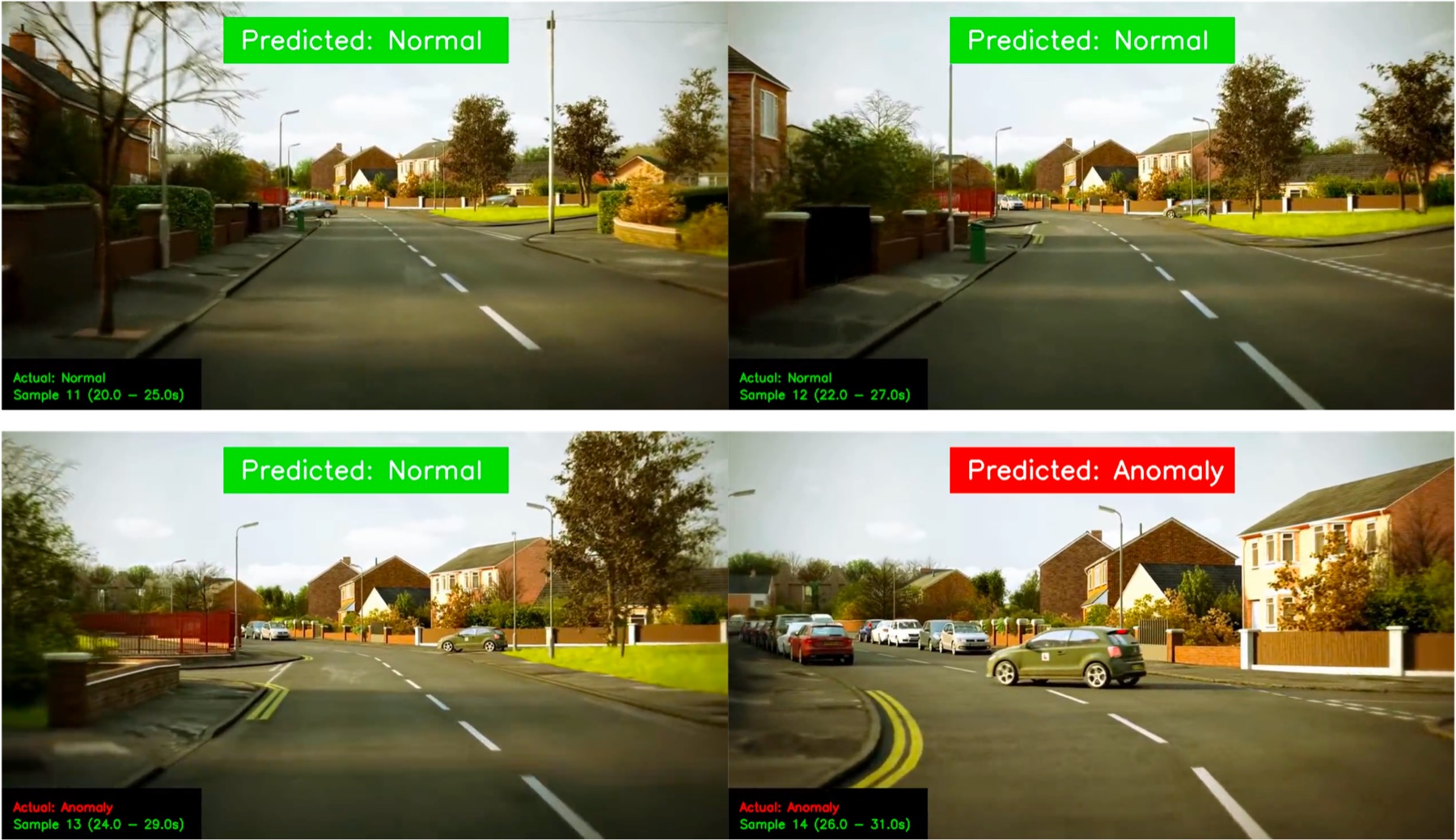}
        \caption{Qualitative results on the Hazard Perception Test Dataset~\cite{theorypass_hazard_perception} using Cosmos-Reason1-7B~\cite{nvidia2025cosmosreason1physicalcommonsense}. Top row (Samples 11--12): correctly classified normal frames. Bottom row (Samples 13--14): a normal frame and a detected anomaly, demonstrating context-aware semantic reasoning.}
        \label{fig:cosmos_output}
    \end{figure}
    
Current AV anomaly detectors are brittle to out-of-distribution objects because they lack semantic and temporal reasoning~\cite{elhafsi2023semantic}. LLM-based approaches recover this reasoning capability but are too slow for on-vehicle deployment, and pipelines that chain a visual parser with a separate LLM compound errors at the vision-language interface~\cite{sinha2024realtimeanomalydetectionreactive, elhafsi2023semantic}. Vision-Language Models (VLMs) address the alignment problem and have been applied to traffic anomaly resolution~\cite{ren2025cot} and multi-agent interaction reasoning~\cite{liu2024reasoning}, yet they inherit the same latency barrier. What is missing is a deployable framework that integrates VLM-level semantic reasoning within the timing budget of a real AV platform---which is precisely what this work builds and benchmarks.

\begin{figure}[ht!]
        \centering
        \includegraphics[width=\linewidth]{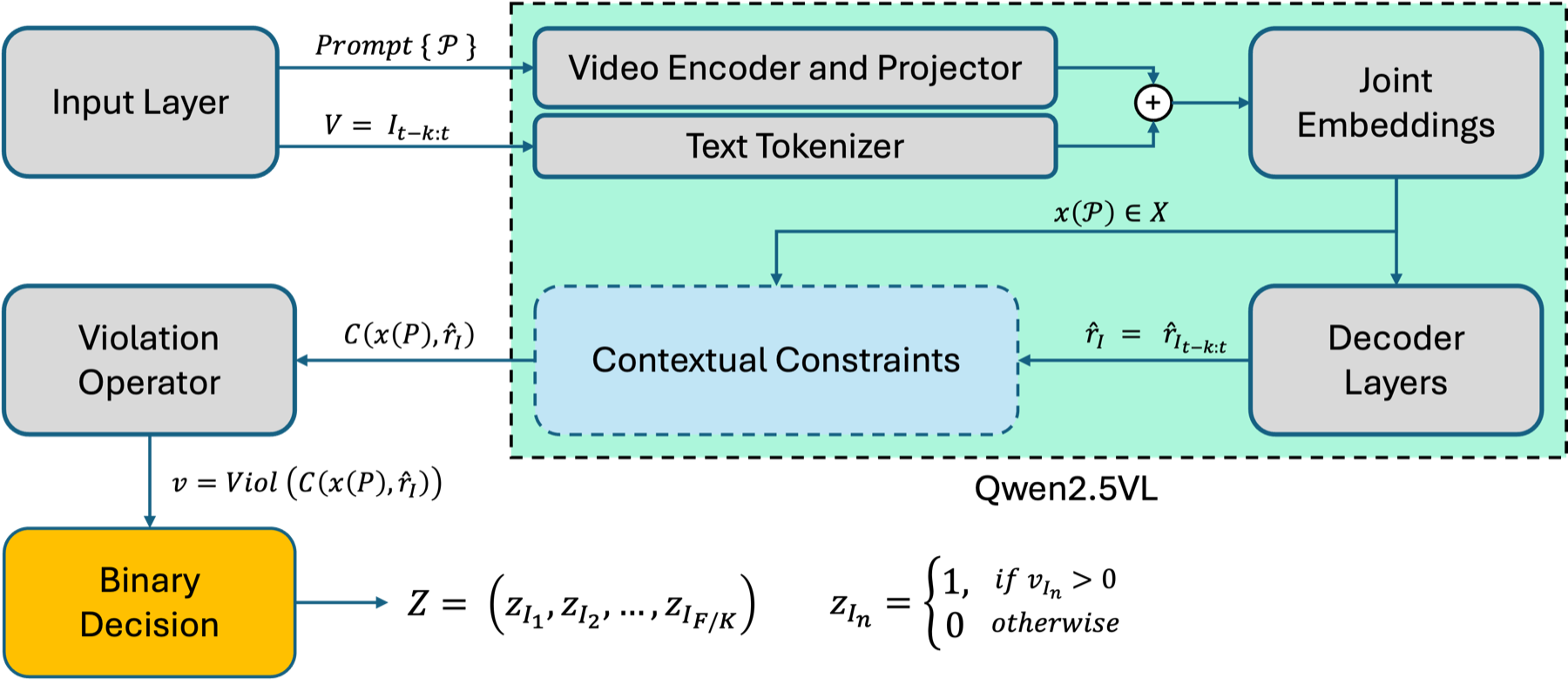}
        \caption{Semantic observer layer architecture. The VLM observer runs at 1--2\,Hz alongside the primary AV control loop, processing temporal windows of RGB frames with a structured prompt. Upon detecting a high-confidence semantic constraint violation, it triggers a fail-safe handoff. Visual tokens from Cosmos-Reason1-7B are projected into the language embedding space and evaluated against context-conditioned semantic constraints to produce a binary \{Normal, Anomaly\} decision.}
        \label{fig:overview}
    \end{figure}

We address these limitations by studying a unique integration of quantized VLMs and attention-efficient inference kernels targeted at an embodied-AI autonomous vehicle test-bed at NYU. The proposed architecture introduces an \textbf{observer-level semantic monitor} that operates alongside—not inside—the primary AV control loop. As illustrated in Fig.~\ref{fig:overview}, this observer layer sits between the regular autonomy stack and the fail-safe stack, acting as an \emph{orchestrator}: running at 1--2\,Hz, continuously monitoring for semantic edge cases the primary stack cannot reason about, and triggering a graceful handoff to the fail-safe stack upon high-confidence anomaly detection. Because the observer is not in the critical control path, its 500 \, ms inference budget is acceptable—semantic anomalies evolve over seconds, not milliseconds—and high precision (minimizing spurious fail-safe activations) is the constraint of the operation design.

This paper proposes and formalizes the \textbf{semantic observer} as a distinct autonomy layer—running at 1--2\,Hz between the regular autonomy stack and the fail-safe stack—responsible for detecting semantic edge cases and orchestrating fail-safe handoffs, decoupling high-level scene reasoning from the primary control loop. We provide the first systematic evaluation of Cosmos-Reason1-7B as a semantic observer, characterizing accuracy, latency, and quantization behavior across static image and video conditions on three public datasets. A key negative finding is that NF4 4-bit quantization causes catastrophic recall collapse (10.6\%) under video inference while performing competitively on static images, establishing a hard per-modality deployment constraint with actionable quantization selection guidance for safety-critical AV systems.

\section{PROBLEM FORMULATION}
\label{sec:problem_formulation}
    To define anomalies in a mathematically principled manner, we formalize the system in terms of the tasks performed by a VLM to produce external anomaly decisions. The system operates on forward-facing RGB video frames from a vehicle in motion, processed in temporally batched windows, together with a structured text prompt that describes semantic constraints and the desired output format.
    
    \medskip
    The prompt is designed to elicit context-aware reasoning over scene elements and their consistency with nominal driving behavior. The output is a binary class $\text{\{normal}, \text{anomaly\}}$. This is the chosen format because we prefer an incremental approach rather than quantifying a severity score. For future implementations and adaptations such as anomaly mitigation, severity/risk scores would help shape control strategies.

    \subsection{\textbf{System Definition}}
        We represent the external anomaly detection system as a tuple
            \begin{equation}
            T_{\mathrm{ext}} = \{S, O_{\mathrm{rgb}}, X, U, Z, \Phi, \varepsilon\},
            \end{equation}
        where \(S\) denotes the ego-vehicle state space, \(O_{\mathrm{rgb}}\) denotes RGB visual observations, \(X\) denotes contextual information, \(U\) denotes predictive uncertainty, \(Z\) denotes the binary external anomaly output, \(\Phi\) denotes explanation artifacts, and \(\varepsilon\) denotes system effectiveness metrics.

    \subsection{\textbf{Video Representation}}
    
        Let \(V\) be an RGB video consisting of \(F\) frames:
            \begin{equation}
            V = \{ I_i \}_{i=1}^{F}, \quad I_i \in O_{\mathrm{rgb}},
            \end{equation}
        where \(I_i\) is the \(i\)-th frame of the video. Each frame is temporally aligned with an ego state $s_i \in S$ and contextual information $x\in X,$ which may include road geometry, intersection topology, traffic rules, and route intent.
    
    \subsection{\textbf{Semantic Scene Interpretation}}
    
        Given a temporal window of frames \(I_{i-k:i}\), a vision-language model maps the visual input and a fixed prompt \(\mathcal{P}\) to a semantic scene representation
            \begin{equation}
            f_{\mathrm{VLM}} : O_{\mathrm{rgb}}^{k+1} \times \mathcal{P} \rightarrow \mathcal{R}
            \end{equation}
            \begin{equation}
            \hat{r}_{I_{i-k:i}} = f_{\mathrm{VLM}}(I_{i-k:i}, \mathcal{P}),
            \end{equation}
        where \(\hat{r}_{I_{i-k:i}} \in \mathcal{R}\) encodes inferred objects, agents, traffic signals, and scene attributes. The prompt \(\mathcal{P}\) conditions the model to reason about safety-relevant and context-dependent anomalies.
    
    \subsection{\textbf{Contextual Semantic Constraints}}
    
        Let $C(x(\mathcal{P}),\hat{r}_{I_{i-k:i}})$ denote a set of semantic constraints defining nominal external behavior under context \(x \triangleq x(\mathcal{P}) \in X\), as interpreted through the prompt \(\mathcal{P}\). These constraints encode expectations such as:
        \begin{itemize}
            \item absence of unexpected objects in drivable regions,
            \item consistency of traffic-light states with intersection semantics,
            \item static behavior of road signs and infrastructure,
            \item compatibility of observed agent behavior with traffic rules.
        \end{itemize}
        
    \subsection{\textbf{Constraint Violation and Binary Anomaly Output}}
    
        We define a semantic violation operator as Equation \ref{eq_5} which evaluates whether the inferred semantics \(\hat{r}_{I_{i-k:i}}\) violate any constraint in $C(x(\mathcal{P}),\hat{r}_{I_{i-k:i}})$. The operator returns a non-negative value indicating the presence of a violation. The binary external anomaly output is defined by Equation \ref{eq_6}.

        \begin{equation} \label{eq_5}
            v_{I_{i-k:i}} = \mathrm{Viol}\big(C(x(\mathcal{P}),\hat{r}_{I_{i-k:i}}) \big),
        \end{equation}
        
        \begin{equation} \label{eq_6}
            z_I =
            \begin{cases}
            1, & \text{if } v_I > 0, \\
            0, & \text{otherwise}.
            \end{cases}
        \end{equation}

        \begin{equation} \label{eq_7}
        Z = (z_{I_1}, z_{I_2}, \dots, z_{I_{\frac{F}k}}), \quad Z_I \in \{0,1\},
        \end{equation}
        Collectively, the system output over the video is defined by Equation \ref{eq_7} where each \(z_{I_n}\) indicates whether window \(I_{i-k:i}\) is classified as externally anomalous. External anomalies are defined as violations of prompt-conditioned semantic constraints rather than probabilistic deviations.

\section{METHODOLOGY}
\label{sec:methodology}



    \subsection{System Architecture}
        Our anomaly detection framework is built upon Cosmos-Reason1-7B\cite{nvidia2025cosmosreason1physicalcommonsense}, a robotics-specialized vision-language model developed by NVIDIA. Architecturally, Cosmos-Reason1-7B retains the full multimodal structure of Qwen2.5-VL \cite{bai2025qwen25vltechnicalreport}, consisting of (i) a ViT-based vision encoder, (ii) an MLP-based vision–language projector, and (iii) a decoder-only transformer backbone. Cosmos is initialized from Qwen2.5-VL and subsequently fine-tuned on robotics and embodied reasoning data, while preserving the underlying architectural design.

        \begin{figure}[ht!]
            \centering
            \includegraphics[width=\linewidth]{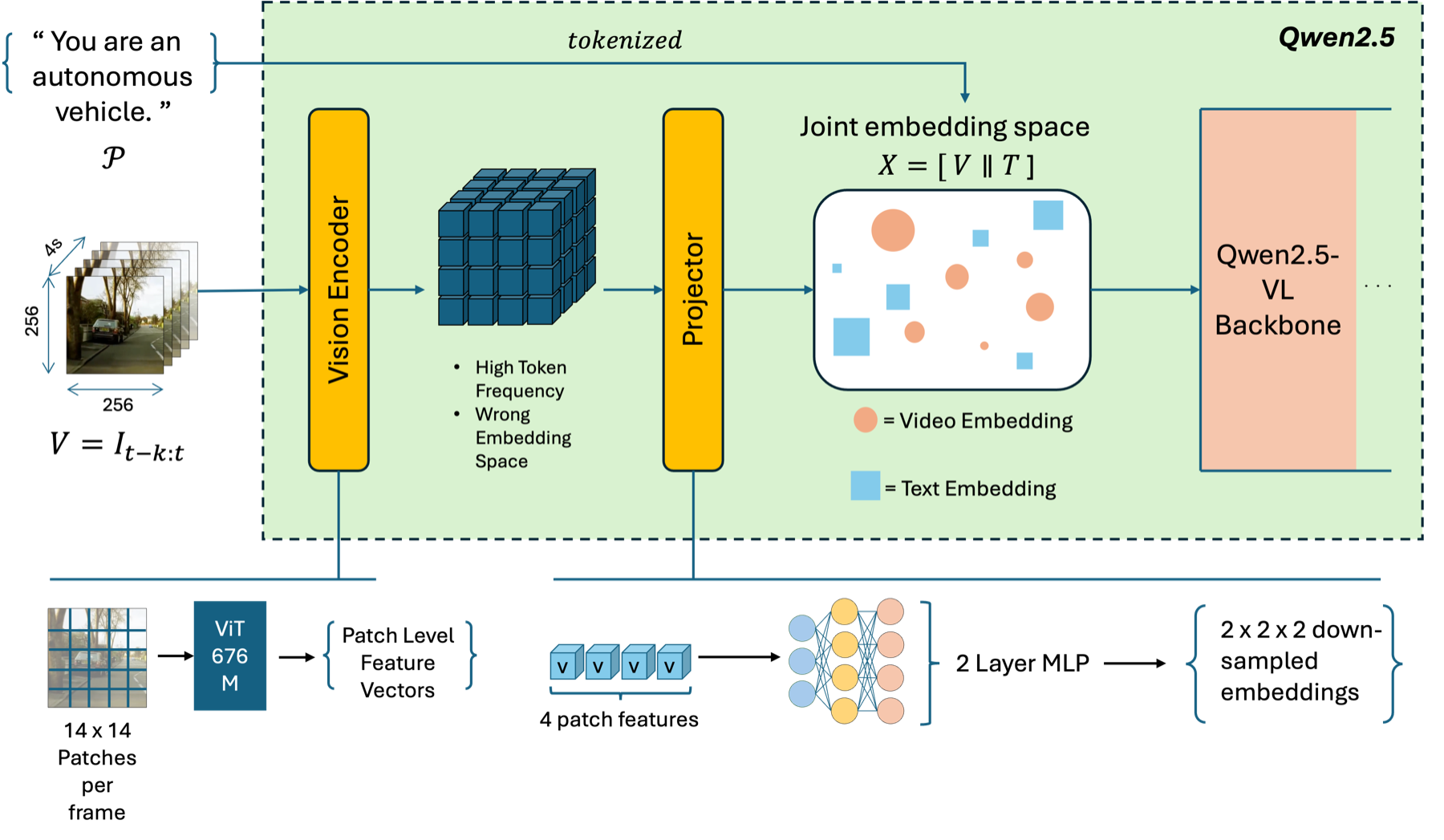}
            \caption{High-level architecture of Cosmos-Reason1-7B for anomaly detection. Visual features from the vision encoder are projected into the language embedding space and jointly processed with prompt tokens by a decoder-only transformer backbone (see Fig.~4 for block details).}
            \label{fig:system_architecture}
        \end{figure}
        
        \medskip
        To enable joint reasoning with language tokens, spatially adjacent patch features are grouped and passed through a two-layer MLP merger (projector), which compresses and maps vision features into the language embedding dimension. The resulting visual tokens are concatenated with prompt tokens to form a unified sequence:
        $$X = [V \;\|\; T].$$
        
        \medskip
        Given a temporally batched RGB window $I_{t-k:t}$ and structured prompt $\mathcal{P}$, visual frames are first processed by the Qwen2.5-VL vision encoder. The encoder splits each frame into patches (stride 14) and applies window-based attention with 2D positional encoding to generate spatial feature tokens. For video inputs, frames are independently encoded and concatenated, with Multimodal Rotary Position Embedding (MRoPE) aligning temporal indices to absolute time to preserve temporal ordering.

        \medskip
        This sequence is processed by a decoder-only transformer backbone inherited from Qwen2.5-VL {~\cite{bai2025qwen25vltechnicalreport}}. Each transformer block consists of RMSNorm, causal self-attention, and a SwiGLU-based feedforward network with residual connections. The backbone outputs contextualized hidden states, which are projected through the language modeling head to produce vocabulary logits. Constrained decoding yields a short textual output (“Normal” or “Anomaly”), which is mapped to a binary decision $z_t \in \{0,1\}$.

    \subsection{Temporal Reasoning over Video Windows}

        External anomalies often depend on short-term dynamics rather than single-frame cues. We therefore perform inference over sliding windows of $k$ consecutive frames,
        \[
        I_{t-k:t},
        \]
        allowing the VLM to reason over local temporal context. In all video experiments (Experiment III) we use $k=5$ frames sampled from 5-second windows at 1\,fps, with a 2-second stride between windows. This covers the typical duration of emerging hazard events (e.g., a vehicle beginning to drift or an obstacle entering the lane) while keeping the token count per inference call bounded. At the 1--2\,Hz observer rate, a 5-second window corresponds to 5--10 observer cycles, providing sufficient overlap for consistent detection without redundant computation. A systematic ablation across window sizes ($k \in \{1, 3, 5, 10\}$) is left for future work.
    
    \subsection{Prompt Engineering and Optimization}
        Prompt design directly influences both reasoning robustness and computational latency. Since decoder-based inference scales approximately linearly with token count, minimizing prompt length is critical for real-time deployment. Let $L(P)$ denote the token length of prompt $P$. Reducing $L(P)$ decreases decoding steps and memory overhead while preserving semantic conditioning.
        
        The final prompt encodes scene understanding assumptions, safety constraints, and strict output formatting in a compact form. Open ended explanations are removed to suppress verbose generation and ensure deterministic outputs. A representative structure is:
        
        \begin{tcolorbox}[colback=gray!10!white, colframe=black, title=LLM Hazard Detection and Communication Prompt, sharp corners=south, boxrule=0.5mm]
        \textit{"You are an autonomous vehicle. Analyze the scene and determine whether it violates normal driving expectations. Output only one word: `Anomaly' or `Normal'."}
        \end{tcolorbox}
        
        Empirically, overly descriptive prompts increased latency without improving detection performance, whereas minimal structured prompts reduced inference time while preserving robustness. The final verbose prompt achieves a balance between semantic expressiveness and computational efficiency by enumerating damage types explicitly (longitudinal cracks, potholes, alligator cracking), providing severity thresholds, and requiring structured \texttt{<think>/<answer>} XML output—scaffolding shown in Experiment II to be essential for NF4 inference quality.

    \subsection{Token Budgeting}
        VLM decoding is autoregressive, and unconstrained generation introduces latency variability. To ensure deterministic behavior, we explicitly constrain generated tokens as
        \[
        T_{\text{gen}} \leq T_{\text{max}}.
        \]
        Original outputs such as ``Classification: Anomaly'' required 4-7 tokens, thereby reducing generation length and eliminating spillover text. This restriction reduces decoding overhead and stabilizes inference latency, contributing significantly to real time performance.

    \subsection{NVFP4 Quantization and FlashAttention2 Acceleration}

    \begin{figure}[ht!]
        \centering
        \includegraphics[width=\linewidth]{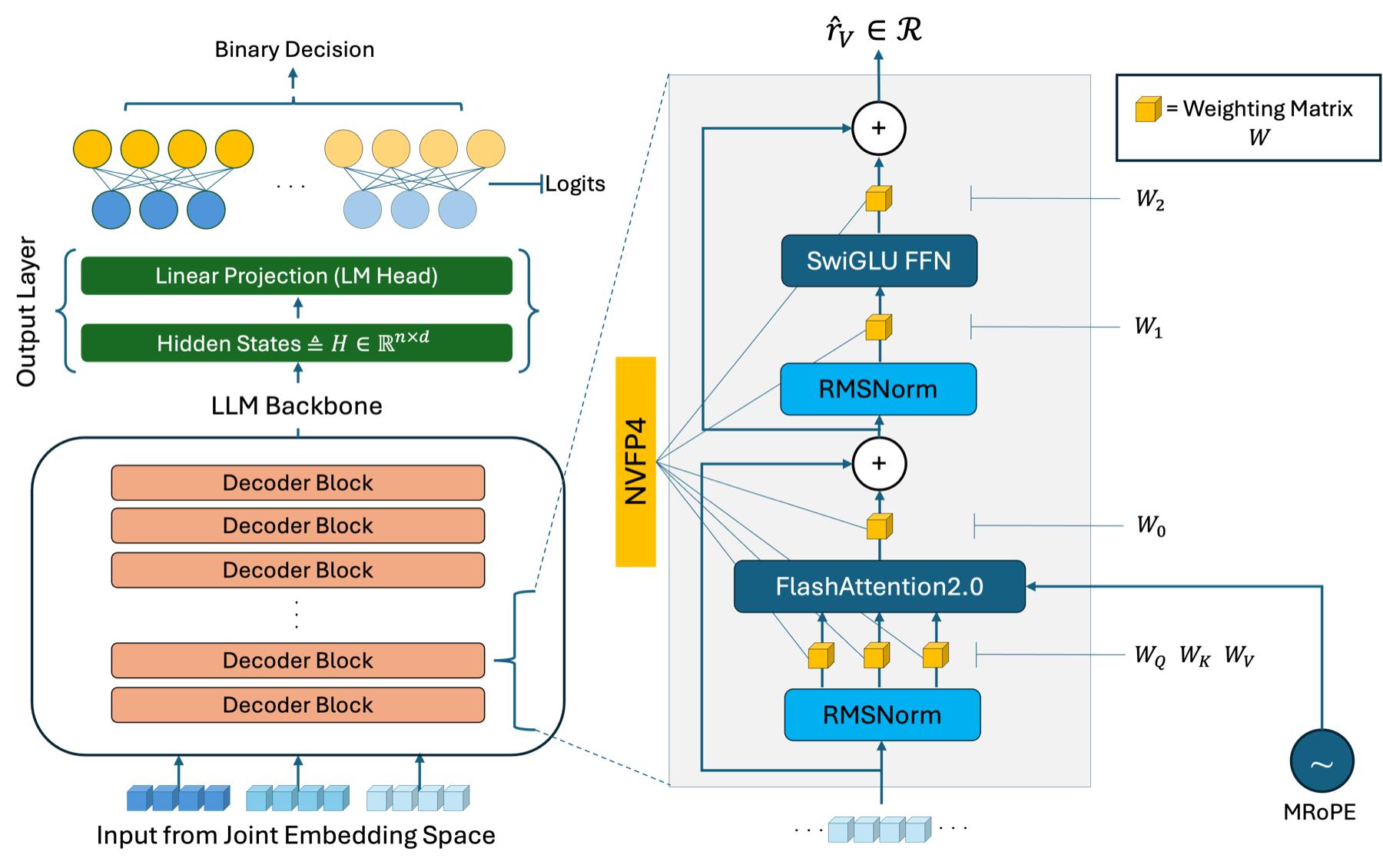}
        \caption{Architecture used in Cosmos-Reason1-7B. Visual tokens extracted by the Qwen2.5-VL vision encoder are projected into the language embedding space via a two-layer MLP merger and concatenated with prompt tokens. NVFP4 quantization is applied to the backbone weight matrices, and FlashAttention2 accelerates attention computation. Cosmos-Reason1-7B retains the Qwen2.5-VL architecture and is further fine-tuned on robotics and embodied reasoning data for physical AI tasks.}
        \label{fig:decoder_arch}
    \end{figure}
    
    To enable deployment under on-vehicle computational constraints, we optimize the decoder-only transformer backbone of Cosmos-Reason1-7B through low-bit weight quantization and memory-efficient attention computation.

    \medskip
    \paragraph{NVFP4 Weight Quantization.}
    4-bit NVFP4 quantization is applied to the linear weight matrices within each transformer block of the language backbone. Specifically, quantization is applied to the projection matrices of the self-attention mechanism,
    \[
    W_Q,\; W_K,\; W_V,\; W_O,
    \]
    as well as to the feedforward network (FFN) weights,
    \[
    W_1,\; W_2,
    \]
    corresponding to the SwiGLU-based MLP layers. 
    
    \medskip
    Let $W \in \mathbb{R}^{m \times n}$ denote a full-precision weight matrix. Under NVFP4 quantization, $W$ is stored in 4-bit precision with per-channel scaling, reducing memory from $16mn$ bits (FP16) to approximately $4mn$ bits, yielding a theoretical $4\times$ reduction in parameter storage. 
    
    \medskip
    Quantization is applied only to the transformer backbone. The vision encoder and vision–language projector remain in higher precision to preserve spatial feature fidelity and multimodal alignment stability.

    \medskip
    \paragraph{FlashAttention2 Kernel Acceleration.}
    Within each transformer block, causal self-attention computes
    \[
    \mathrm{Attn}(Q,K,V) = \mathrm{softmax}\!\left({QK^\top}\right)V,
    \]
    where the quadratic token interaction $QK^\top \in \mathbb{R}^{n \times n}$ introduces significant memory traffic for long multimodal sequences.
    
    \medskip    
    FlashAttention2 \cite{dao2024flashattention2} replaces the standard attention implementation with a tiled, memory-efficient kernel that computes attention without materializing the full $n \times n$ matrix in high-bandwidth memory. Instead, attention is computed in streaming blocks using on-chip SRAM, reducing memory IO while preserving numerically exact attention outputs (up to floating-point precision). Key improvements over the original FlashAttention include reduced non-matmul FLOPs in the forward pass, query-partitioned warp splitting in the backward pass, and sequence-length parallelization that raises GPU SM occupancy from $<$50\% to $>$70\% for long-context inference.

    \medskip
    \paragraph{Combined Effect.}
    NVFP4 reduces the memory footprint and matrix multiplication cost of the transformer’s linear operators, while FlashAttention2 reduces the memory bandwidth overhead of self-attention. Since transformer inference cost is dominated by (i) large GEMM operations and (ii) quadratic attention memory movement, these optimizations address both primary bottlenecks. The total runtime is expressed as
        \[
        T_{\text{total}} = T_{\text{sense}} + T_{\text{preprocess}} + T_{\text{infer}} + T_{\text{post}},
        \]
    where $T_{\text{infer}}$ is dominated by transformer computation. The unoptimized FP16 baseline---Cosmos-Reason1-7B in standard PyTorch with no quantization and no FlashAttention2 kernel on the same RTX\,5090---requires approximately $\sim$25\,s per frame, consistent with published throughput figures for 7B-scale VLMs in this configuration. Under the combined NVFP4+FlashAttention2 configuration this reduces to $\sim$500\,ms, a $\sim$50$\times$ speedup. The quantization variants in Table~\ref{tab:video_quantization} all employ FlashAttention2 and differ only in weight precision; their mutual latency differences therefore reflect quantization-specific overhead rather than the full optimization stack. Notably, INT8 is slower than BF16 (0.787\,s vs.\ 0.485\,s) because modern Tensor Cores execute BF16 GEMM at near-theoretical peak throughput, whereas INT8 requires per-token activation scaling and dequantization steps whose overhead outweighs the arithmetic savings at the short sequence lengths encountered here. An overview of the complete system is illustrated in Figure~\ref{fig:decoder_arch}.

\section{RESULTS}
\label{sec:results}
        \subsection{Experiment I: Paradigm Comparison --- Statistical vs.\ Semantic Anomaly Detection}
    
        This experiment is a \emph{paradigm illustration}, not a performance comparison. FCDD and Cosmos-Reason1-7B are not competing solutions to the same task: FCDD learns pixel-level distributional deviations from nominal data, while Cosmos detects context-conditioned semantic constraint violations (Sec.~\ref{sec:problem_formulation}). Accordingly, they are evaluated on fundamentally different metrics (ROC-AUC vs.\ F1/Precision/Recall) that reflect their respective objectives.

        An important methodological caveat is that the normal class (Cityscapes urban road crops) and the anomalous class (RDD2022 road damage) originate from different acquisition domains and camera systems. FCDD's near-perfect ROC-AUC ($\approx$1.0) is therefore best interpreted as a \emph{domain-shift artifact}: the model learns to separate acquisition conditions rather than abstract damage semantics. Rather than evaluating a domain-adapted FCDD baseline---which would require dataset alignment beyond our scope---we use the experiment to contrast \emph{what each model can say} about the same anomalous image: FCDD produces an anomaly heat map without any label; Cosmos produces an actionable semantic classification. This qualitative gap, illustrated in Figs.~\ref{fig:fcdd_heatmaps} and~\ref{fig:cosmos_output}, is the central motivation for the observer-layer design.
    
        \medskip
        \subsubsection{FCDD Training and Dataset Configuration}
        
\begin{figure}[ht!]
            \centering
            \includegraphics[width=\linewidth]{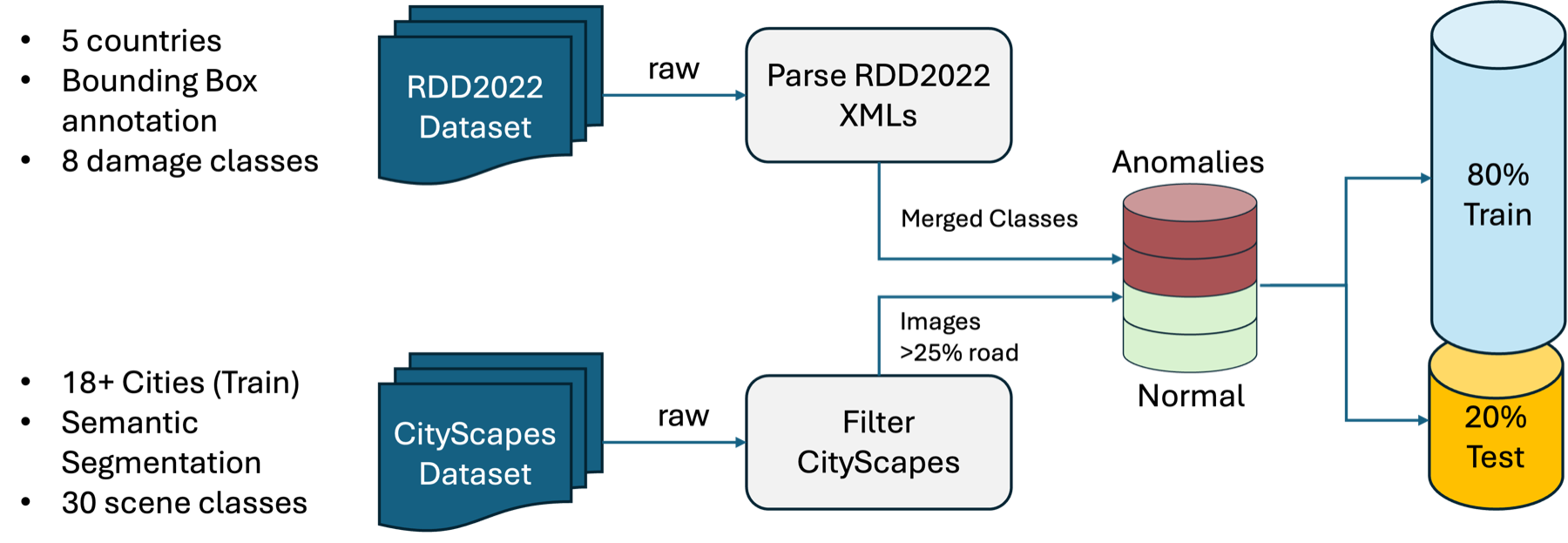}
            \caption{Dataset orchestration for FCDD training. RDD2022\cite{arya2024rdd2022} images (all damage types merged) serve as the anomalous class, while Cityscapes\cite{cordts2016cityscapes} images filtered for $\geq$25\% road coverage provide the normal class. An 80/20 train-test split yields 31,386 training samples (2,598 normal, 28,788 anomalous) and 7,643 test samples (447 normal, 7,196 anomalous).}
            \label{fig:fcdd_dataset_orchestration}
        \end{figure}

        FCDD \cite{liznerski2021explainable} operates as a static image anomaly detector, learning pixel-level deviations from nominal road surfaces \textbf{without temporal reasoning} capabilities. We trained FCDD on a binary classification task: clean roads (normal) versus damaged roads (anomalous). Figure~\ref{fig:fcdd_dataset_orchestration} illustrates our dataset preparation pipeline.
    
\begin{figure}[ht!]
            \centering
            \begin{subfigure}{\linewidth}
                \includegraphics[width=\linewidth]{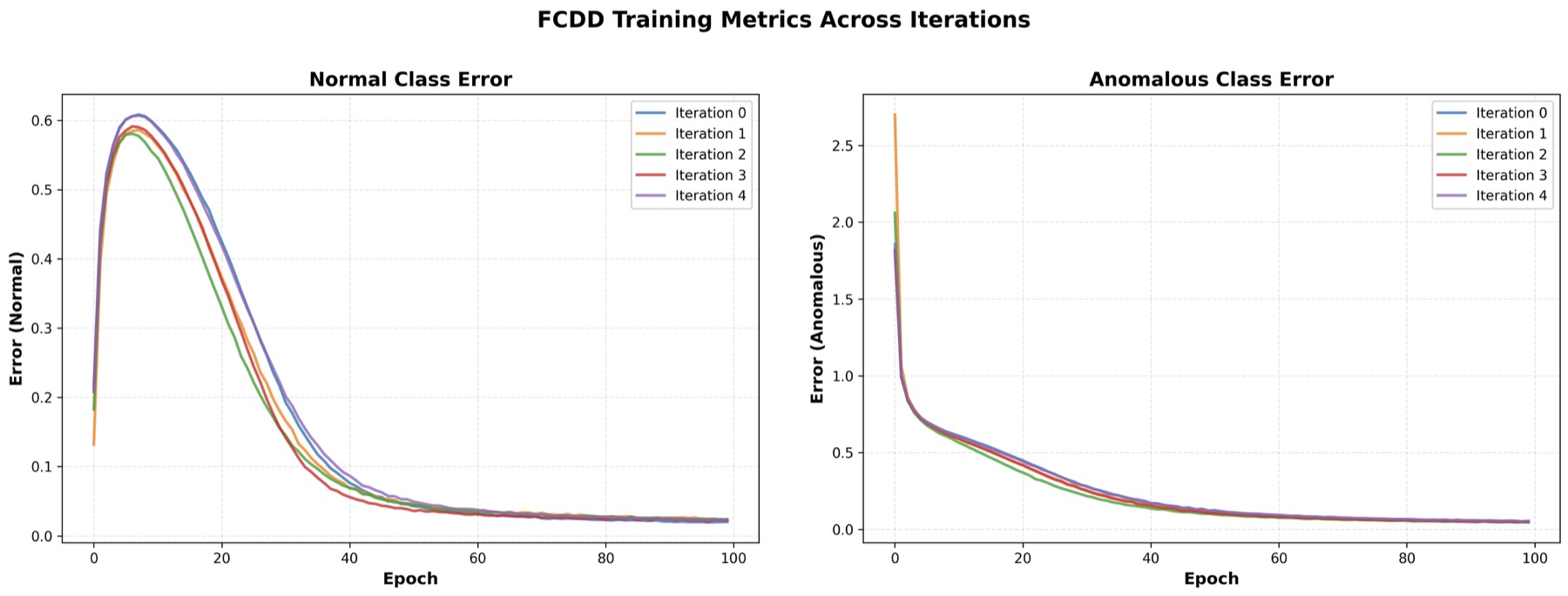}
                \caption{Class-specific error curves}
                \label{fig:fcdd_err_classes}
            \end{subfigure}
            
            \vspace{0.5cm} 
            
            \begin{subfigure}{\linewidth}
                \includegraphics[width=\linewidth]{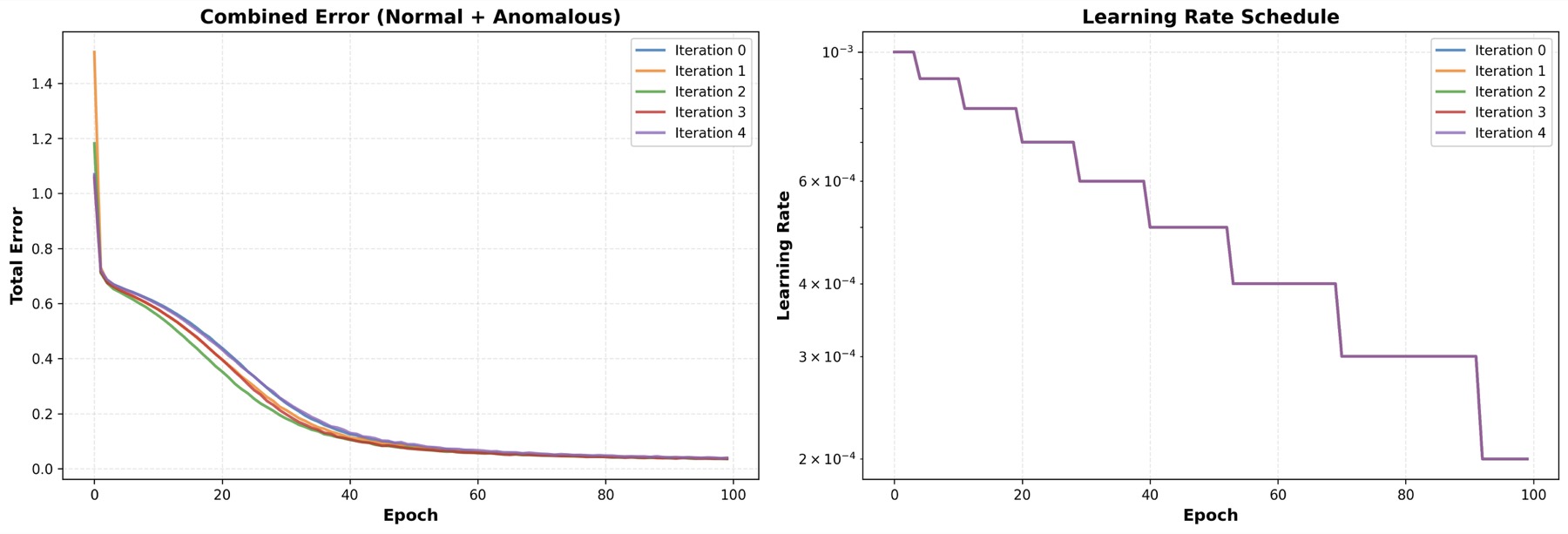}
                \caption{Combined error and learning rate}
                \label{fig:fcdd_err_lr}
            \end{subfigure}
            \caption{FCDD training metrics across 5 iterations. (a) Both class errors converge near 0.0 after 100 epochs with maintained separation. (b) Combined error decreases under step learning rate decay ($10^{-3} \to 2\times10^{-4}$ at epoch 100).}
            \label{fig:fcdd_training_curves}
        \end{figure}

        \medskip
        The model converged after approximately 100 epochs across 5 iterations with different random seeds, achieving near-perfect separation on the test set (ROC-AUC $\approx$ 1.0). Figure~\ref{fig:fcdd_training_curves} shows the training dynamics across iterations.
        
        \medskip
        \subsubsection{Limitations of Static Anomaly Detection}
        
\begin{figure}[ht!]
            \centering
            \begin{subfigure}{\linewidth}
                \includegraphics[width=\linewidth]{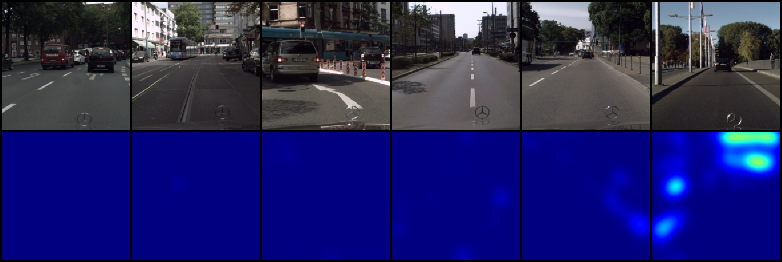}
                \caption{Normal class (Cityscapes)}
                \label{fig:fcdd_normal}
            \end{subfigure}
            \hfill
            \begin{subfigure}{\linewidth}
                \includegraphics[width=\linewidth]{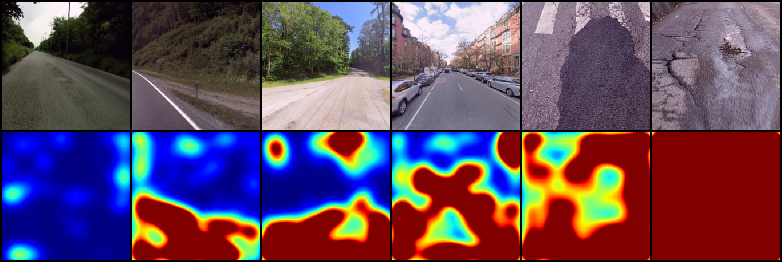}
                \caption{Anomalous class (RDD2022)}
                \label{fig:fcdd_anomalous}
            \end{subfigure}
            \caption{FCDD global heatmaps. Red regions indicate high anomaly scores. (a) Clean roads receive uniformly low scores. (b) Damaged sections trigger pronounced activations.}
            \label{fig:fcdd_heatmaps}
        \end{figure}

        While FCDD successfully learns distributional patterns distinguishing damaged from clean roads, it exhibits fundamental limitations when applied to embodied driving scenarios. Figure~\ref{fig:fcdd_heatmaps} shows representative detections on test samples.
        
        \medskip
        FCDD's limitations stem from three core gaps: (1) \textbf{No temporal reasoning}---each frame is processed independently without motion context, precluding detection of time-evolving hazards (e.g., drifting vehicles); (2) \textbf{No semantic grounding}---pixel-level scores indicate ``unusual texture'' but lack actionable labels (pothole vs.\ shadow vs.\ construction zone), preventing mitigation planning; (3) \textbf{Domain shift}---because the normal class (Cityscapes) and anomalous class (RDD2022) differ in camera, geography, and acquisition conditions, the near-perfect ROC-AUC reflects cross-domain separation rather than abstract damage reasoning, as evidenced by false positives on lane markings in Fig.~\ref{fig:fcdd_heatmaps}. In contrast, Cosmos classifies the same images with natural-language rationale, providing the semantic label and recommended action that are inaccessible to any pixel-level detector.
        
            \subsubsection{Paradigm Contrast: What Each Model Can and Cannot Provide}
                Table~\ref{tab:fcdd_vs_cosmos} summarizes the capabilities of each model on the RDD2022 test set. The differing metrics reflect each model's design objective and should not be interpreted as one model outperforming the other.
            
                \medskip
                FCDD achieves near-perfect ROC-AUC (1.00), demonstrating that pixel-level distributional modeling can reliably separate clean urban roads (Cityscapes) from damaged ones (RDD2022). However, this strong result is specific to a fixed domain split and does not imply that FCDD can generalize to the semantic edge cases that motivate this work.
                
                \begin{table}[t]
                        \centering
                        \caption{Paradigm comparison of FCDD and Cosmos-Reason1-7B. The two models solve fundamentally different problems and are evaluated on different metrics by design: FCDD optimizes distributional separation (ROC-AUC) while Cosmos produces actionable semantic classifications (F1/Precision/Recall). Direct metric comparison is intentionally omitted. Note that the models are also evaluated on different test sets: FCDD on a held-out split of RDD2022+Cityscapes (7,643 images); Cosmos on the same RDD2022+Cityscapes split used in Experiment II (2,162 images). *ROC-AUC measures ranking quality; F1/Precision/Recall measure classification at a fixed threshold.}
                        \label{tab:fcdd_vs_cosmos}
                        \begin{threeparttable}
                        \begin{tabular}{lcccc}
                        \toprule
                        Model & ROC-AUC$^*$ $\uparrow$ & F1 $\uparrow$ & Precision $\uparrow$ & Recall $\uparrow$ \\
                        \midrule
                        FCDD & 1.00 & — & — & — \\
                        Cosmos (7B, NF4) & — & 0.60 & 0.83 & 0.47 \\
                        \bottomrule
                        \end{tabular}
                        \begin{tablenotes}
                            \footnotesize
                            \item ROC-AUC refers to Receiver operating characteristic - Area under curve
                        \end{tablenotes}
                        \end{threeparttable}
                    \end{table}
                    Cosmos operates under a fundamentally different paradigm: it produces semantic classifications with natural language reasoning, enabling actionable labels (e.g., \textit{pothole}, \textit{crack}, \textit{construction zone}) and explanations that FCDD cannot generate by design. Its F1 (0.60) reflects zero-shot classification across multiple damage categories without any task-specific training. High precision (0.83) indicates reliable predictions when damage is flagged, while lower recall (0.47) reflects conservative behavior under uncertainty. Critically, Cosmos provides \emph{what} the anomaly is and \emph{what to do}—information that is inaccessible to any pixel-level statistical detector.

\subsection{Experiment II: Feasibility of VLM Observers Under Quantization and Prompt Constraints}
            We assess Cosmos-Reason1-7B feasibility as an observer-level semantic monitor by varying quantization (INT8 vs.\ NF4) and prompt verbosity (Verbose, Pruned, Minimal) on the RDD2022+Cityscapes test set (2,162 images; 79\% anomalous). F1 is the primary metric given class imbalance.
            
            \begin{table}[t]
                \centering
                \caption{Cosmos-Reason1-7B performance across quantization and prompt configurations. Dataset imbalance: 79\% anomalous, 21\% normal. F1 score is the primary metric due to class imbalance.}
                \label{tab:cosmos_quantization_prompt}
                \begin{threeparttable}
                \begin{tabular}{llcccc}
                \toprule
                \textbf{Quantization} & \textbf{Prompt} & \textbf{Prec.$\uparrow$} & \textbf{Rec.$\uparrow$} & \textbf{F1$\uparrow$} & \textbf{Lat.(s)$\downarrow$} \\
                \midrule
                \multirow{3}{*}{INT8} 
                    & Verbose       & \textbf{84.1\%} & 45.1\% & 58.7\% & 1.33 \\
                    & Pruned        & 53.9\% & 12.5\% & 20.3\% & 1.37 \\
                    & Minimal   & —  & —  & 0.0\%  & 1.36 \\
                \midrule
                \multirow{3}{*}{NF4} 
                    & Verbose       & 82.8\% & \textbf{47.0\%} & \textbf{60.0\%} & \textbf{0.80} \\
                    & Pruned        & \textbf{98.0\%} & 5.7\% & 10.8\% & 0.85 \\
                    & Minimal   & —  & — & 0.0\% & 0.84 \\
                \bottomrule
                \end{tabular}
                \begin{tablenotes}
                    \footnotesize
                    \item Minimal prompts produced 100\% unparseable outputs (lack of XML structure).
                    \item NF4 provides 1.6$\times$ speedup over INT8 (0.80s vs 1.33s per image).
                \end{tablenotes}
                \end{threeparttable}
            \end{table}
        
            \begin{table}[t]
                \centering
                \caption{Confusion matrices for top-performing Cosmos configurations. Positive class = Anomaly (road damage). NF4+Verbose is the optimal static image configuration: highest F1 (60.0\%), best recall (47.0\%), and fastest inference, demonstrating that aggressive quantization succeeds only when paired with verbose structured prompts.}
                \label{tab:cosmos_confusion}
                \small
                \begin{threeparttable}
                \begin{tabular}{l>{\columncolor{blue!10}}ccc}
                \toprule
                & \textbf{NF4+Verbose} & \textbf{INT8+Verbose} & \textbf{INT8+Pruned} \\
                \midrule
                TP & 806 & 773 & 215 \\
                FN & 909 & 942 & 1500 \\
                FP & 168 & 146 & 184 \\
                TN & 279 & 301 & 263 \\
                \midrule
                \rowcolor{blue!10}F1 Score & 60.0\% & 58.7\% & 20.3\% \\
                Precision & 82.8\% & 84.1\% & 53.9\% \\
                Recall & 47.0\% & 45.1\% & 12.5\% \\
                \bottomrule
                \end{tabular}
                \begin{tablenotes}
                    \footnotesize
                    \item TP = True Positives, TN = True Negatives
                    \item FP = False Positives, FN = False Negatives
                \end{tablenotes}
                \end{threeparttable}
            \end{table}
            
            NF4 dominates INT8 on all metrics (F1: 60.0\% vs.\ 58.7\%, recall: 47.0\% vs.\ 45.1\%) with 40\% faster inference (0.80\,s vs.\ 1.33\,s). Verbose prompts are non-negotiable: F1 collapses to 0\% under minimal prompting. The best configuration (NF4+Verbose) achieves 82.8\% precision and 47.0\% recall—a favorable operating point for the observer role, where spurious fail-safe triggers are more disruptive than missed detections handled by the primary stack. Improving recall is the primary remaining challenge addressed in Sec.~\ref{sec:conclusion}.
        
        \subsection{Experiment III: Temporal Window Analysis and the NF4 Collapse Finding}
    
            We evaluate Cosmos-Reason1-7B on sliding temporal windows of real driving video—the natural operating mode of an observer-level module. This experiment yields a critical negative finding: \textbf{NF4 quantization causes catastrophic recall collapse under video constraints}, which must be treated as a hard deployment constraint for any observer implementation.
            
            \textbf{Experimental Setup.} We process 224 video clips from the Hazard Perception Test Dataset~\cite{theorypass_hazard_perception} using 5-second temporal windows ($k=5$ frames at 1\,fps) with 2-second stride. Each window is analyzed by Cosmos with a fixed-length prompt (max 3 output tokens: ``Anomaly'', ``Normal'', or ``Unknown''). We compare three quantization configurations---BF16 (baseline), INT8, and NF4---on an NVIDIA RTX 5090 (32\,GB VRAM, 108 TFLOPS, CUDA 13.0). This GPU was released in early 2025 and may not be broadly available; latency figures should be interpreted relative to the ratios between configurations (BF16:INT8:NF4 $\approx$ 1:1.6:0.9) rather than as absolute numbers, since relative ordering is expected to hold across comparable Tensor Core architectures (e.g., RTX 4090, A100). Results are presented in Table~\ref{tab:video_quantization}.
            
            \begin{table}[t]
                \centering
                \caption{Cosmos-Reason1-7B video stream performance across quantizations. Hazard Perception Test Dataset (224 clips, 5s windows, 2s stride). BF16 is the optimal configuration: balances recall (77.3\%), F1 (50.8\%), and latency (0.485s), while NF4's speed advantage is negated by catastrophic recall collapse (10.6\%).}
                \label{tab:video_quantization}
                \begin{tabular}{lcccccc}
                \toprule
                \textbf{Quantization.} & \textbf{TP} & \textbf{TN} & \textbf{FP} & \textbf{FN} & \textbf{Precision$\uparrow$} & \textbf{Recall.$\uparrow$} \\
                \midrule
                \rowcolor{blue!10} BF16    & 51 & 96  & 84  & 15 & 37.8\% & \textbf{77.3\%} \\
                INT8    & 50 & 99  & 81  & 16 & 38.2\% & 75.8\% \\
                \rowcolor{blue!10} NF4     & 7  & 162 & 18  & 59 & 28.0\% & 10.6\% \\
                \midrule
                \textbf{Quantization} & \multicolumn{2}{c}{\textbf{F1$\uparrow$}} & \multicolumn{2}{c}{\textbf{Acc.$\uparrow$}} & \multicolumn{2}{c}{\textbf{Latency (s)$\downarrow$}} \\
                \midrule
                \rowcolor{blue!10} BF16    & \multicolumn{2}{c}{\textbf{50.8\%}} & \multicolumn{2}{c}{59.8\%} & \multicolumn{2}{c}{0.485} \\
                INT8    & \multicolumn{2}{c}{\textbf{50.8\%}} & \multicolumn{2}{c}{\textbf{60.6\%}} & \multicolumn{2}{c}{0.787} \\
                \rowcolor{blue!10}NF4     & \multicolumn{2}{c}{15.4\%} & \multicolumn{2}{c}{68.7\%} & \multicolumn{2}{c}{\textbf{0.436}} \\
                \bottomrule
                \end{tabular}
            \end{table}

\section{ANALYSIS}
        \label{sec:analysis}

        \textbf{Detection Paradigm.} FCDD achieves near-perfect statistical separation (ROC-AUC $\approx$ 1.0) on the RDD2022+Cityscapes split, but this largely reflects the cross-domain gap between the two acquisition conditions (Sec.~\ref{sec:results}~A) and provides no actionable labels---FCDD cannot distinguish a pothole from a shadow or a construction zone. Cosmos-Reason1-7B trades some raw recall for interpretable semantic classifications, answering \emph{what} the anomaly is and \emph{what to do}. For the observer-level conflict-avoidance role, this semantic output is the operative signal for triggering a fail-safe handoff. The best Cosmos static-image configuration (NF4+Verbose) achieves a balanced accuracy of 54.7\% (recall 47.0\%, specificity $279/447=62.4\%$), reflecting the conservative zero-shot operating point; video-mode BF16 achieves 65.3\% balanced accuracy (recall 77.3\%, specificity $96/180=53.3\%$), providing a fairer comparison baseline for future work on more balanced splits.

        \medskip
        \textbf{Quantization and Prompt Design.} Structured verbose prompts are non-negotiable: F1 collapses from 60\% to 0\% under minimal prompts. Within the verbose regime, NF4+Verbose (F1: 60\%, precision: 82.8\%, latency: 0.80\,s) is the recommended static-image observer configuration. Under video constraints, NF4 collapses catastrophically—F1: 15.4\%, recall: 10.6\%, an 89\% false-negative rate on safety-critical hazards—while BF16/INT8 maintain F1 of 50.8\% and recall of $\sim$77\% within the 1--2\,Hz observer budget. \textbf{NF4 must not be used in video observer deployments.} Detailed confusion matrices are in Table~\ref{tab:cosmos_confusion}.

        \medskip
        \textbf{Decision Threshold and Fail-Safe Rate Limiting.} The results in this paper use the argmax (50\% confidence threshold), but deployment requires validation-set calibration to jointly satisfy the ASIL-B precision ($\geq$80\%) and ASIL-D recall ($\geq$90\%) targets. To suppress false activations from transient misclassifications, a temporal debouncing mechanism should require $n_{\min}$ consecutive positive detections within a sliding window before activating the MRM; both the threshold and $n_{\min}$ are deployment-time parameters determined by the platform's FMEA.

        \medskip
        \textbf{Dataset Scope and Generalizability.} All three experiments use datasets collected under limited or controlled conditions. RDD2022 covers road surface damage across four countries but excludes semantic anomaly classes such as misread traffic signals, unexpected pedestrian behavior, or debris other than road damage. The Hazard Perception Test dataset uses scripted UK Theory Test clips rather than naturalistic driving footage, which may understate the false-positive rate in open-domain scenarios. The zero-shot Cosmos results on these datasets should therefore be interpreted as a lower bound on performance for in-domain road-damage anomalies and an untested bound on other semantic anomaly categories. Future work should include evaluation on DoTA~\cite{yao2020unsupervised} or DADA-2000~\cite{fang2021dada} datasets, which contain naturalistic accident-adjacent driving scenarios that better represent the full semantic anomaly space targeted by the observer layer.
       \begin{table}[t!]
            \centering
            \caption{HARA for the VLM observer module (ISO\,26262).}
            \label{tab:hara}
            \small
            \begin{tabular}{p{2.8cm}cp{3.2cm}}
            \toprule
            \textbf{Hazardous Event} & \textbf{ASIL} & \textbf{Safety Goal} \\
            \midrule
            False positive: spurious fail-safe trigger & B & Precision $\geq$80\%; debounce triggers \\
            False negative: undetected hazard & D & Recall $\geq$90\%; redundant detection \\
            Excessive latency & B & Latency $\leq$1\,s; watchdog monitor \\
            NF4 silent recall collapse (video) & D & Prohibit NF4 in video path \\
            \bottomrule
            \end{tabular}
        \end{table}
        \medskip

        \textbf{Hazard Analysis and Risk Assessment (HARA).} Table~\ref{tab:hara} summarizes the HARA under ISO\,26262; safety violations can emerge even from accident-free traffic~\cite{tian2026discovering}, so observer-triggered fail-safe switching must be governed by formal safety goals. The HARA identifies two ASIL-D items: undetected hazards and silent NF4 failure. The current system satisfies the ASIL-B precision goal (82.8\%) but not the ASIL-D recall target ($\geq$90\%), leaving a 13-point gap. The static vs.\ video recall gap (47\% vs.\ 77.3\%) reflects distinct failure modes: static inference lacks surrounding scene context, yielding conservative single-frame judgments, while video inference leverages temporal motion cues to confirm hazard presence. To close this gap we plan (i) rank-16/32 LoRA fine-tuning on 2k--5k labeled frames, (ii) multi-frame logit aggregation to suppress single-frame false negatives, and (iii) confidence-calibrated threshold tuning. Until the recall target is met the observer must not serve as the sole safety layer; the MRM layer of~\cite{arab2024high}---which provides reachability-based collision-free trajectory guarantees compatible with the observer's binary trigger---can be formally integrated only once safety goals are satisfied.
\section{CONCLUSIONS} 
        \label{sec:conclusion}
        This paper presented a pre-deployment feasibility study of quantized VLMs as a semantic observer layer for embodied-AI AV systems, targeting an observer-level orchestrator that minimizes driving hazards by triggering fail-safe handoffs when semantic edge cases are detected. All experiments were conducted on public datasets (RDD2022, Cityscapes, Hazard Perception Test); on-vehicle validation on the NYU AV test-bed is the direct next step. Statistical detectors cannot provide the actionable classifications required for this role; Cosmos-Reason1-7B with NF4+Verbose quantization achieves 0.80\,s inference and 82.8\% precision, satisfying the 1--2\,Hz observer timing budget. A critical negative finding is that NF4 causes catastrophic recall collapse (10.6\%) under video constraints---BF16 or INT8 are the only safe choices, yielding F1 of 50.8\% and recall of $\sim$77\%. The current system meets the ASIL-B precision goal but not the ASIL-D recall target ($\geq$90\%), and must not be deployed as the sole safety layer until that gap is closed through LoRA fine-tuning, multi-frame score aggregation, and calibrated threshold selection. These results establish the feasibility of the architecture and provide a concrete roadmap for safety-compliant integration with a certified MRM layer on the NYU AV platform.

\section*{Acknowledgment}
The authors thank the Department of Mechanical and Aerospace Engineering at the Tandon School of Engineering at New York University for providing computational resources and institutional support. Special thanks to Prof. Katsuo Kurabayashi and Dr. Ray Li for their guidance throughout this project.

\bibliographystyle{ieeetr}
\bibliography{references}

@inproceedings{nwankwo2025envodat,
  title={EnvoDat: A Large-Scale Multisensory Dataset for Robotic Spatial Awareness and Semantic Reasoning in Heterogeneous Environments},
  author={Nwankwo, L. and Ellensohn, B. and Dave, V. and Hofer, P. and others},
  booktitle={IEEE Int. Conf. Robot. Autom. (ICRA)},
  pages={153--160},
  year={2025},
  organization={IEEE}
}

@inproceedings{hanson2022vast,
  title={{VAST}: Visual and Spectral Terrain Classification in Unstructured Multi-Class Environments},
  author={Hanson, N. and Shaham, M. and Erdo{\u{g}}mu{\c{s}}, D. and Padir, T.},
  booktitle={IEEE/RSJ Int. Conf. Intell. Robots Syst. (IROS)},
  pages={3956--3963},
  year={2022},
  organization={IEEE}
}

@article{duan2026causalnav,
  title={{CAUSALNAV}: A Long-Term Embodied Navigation System for Autonomous Mobile Robots in Dynamic Outdoor Scenarios},
  author={Duan, H. and Luo, S. and Deng, Z. and Chen, Y. and others},
  journal={IEEE Robot. Autom. Lett.},
  year={2026},
  publisher={IEEE}
}

@inproceedings{liang2025gnd,
  title={{GND}: Global Navigation Dataset with Multi-Modal Perception and Multi-Category Traversability in Outdoor Campus Environments},
  author={Liang, J. and Das, D. and Song, D. and Shuvo, M. N. H. and others},
  booktitle={IEEE Int. Conf. Robot. Autom. (ICRA)},
  pages={2383--2390},
  year={2025},
  organization={IEEE}
}

@article{tian2026discovering,
  title={Discovering Safety Violations of Decision-Making in Autonomous Driving Systems from Accident-Free Traffic Scenarios},
  author={Tian, H. and Zhou, Y. and Guo, A. and Wu, G. and others},
  journal={ACM Trans. Softw. Eng. Methodol.},
  year={2026},
  publisher={ACM}
}

@article{liu2024reasoning,
  title={Reasoning Multi-Agent Behavioral Topology for Interactive Autonomous Driving},
  author={Liu, H. and Chen, L. and Qiao, Y. and Lv, C. and Li, H.},
  journal={Adv. Neural Inf. Process. Syst.},
  volume={37},
  pages={92605--92637},
  year={2024}
}

@article{arab2024high,
  title={High-Resolution Safety Verification for Evasive Obstacle Avoidance in Autonomous Vehicles},
  author={Arab, A. and Khaleghi, M. and Partovi, A. and Abbaspour, A. and others},
  journal={IEEE Open J. Veh. Technol.},
  volume={6},
  pages={276--287},
  year={2024},
  publisher={IEEE}
}

@inproceedings{vojir2021road,
  title={Road anomaly detection by partial image reconstruction with segmentation coupling},
  author={Vojir, T. and {\v{S}}ipka, T. and Aljundi, R. and Chumerin, N. and others},
  booktitle={IEEE/CVF Int. Conf. Comput. Vis. (ICCV)},
  pages={15651--15660},
  year={2021},
  organization={IEEE}
}

@article{tian2024latency,
  title={Latency-aware road anomaly segmentation in videos: A photorealistic dataset and new metrics},
  author={Tian, B. and Gao, H. and Cui, L. and Zheng, Y. and others},
  journal={arXiv preprint arXiv:2401.04942},
  year={2024}
}

@inproceedings{ZhengVideoBasedTA,
  title={Video-Based Traffic Anomaly Detection with Vision-Language Models: A Survey},
  author={Zheng, Y. and Zhou, W. and Nastic, S. and Wang, C.},
  url={https://api.semanticscholar.org/CorpusID:280315915}
}

@article{elhafsi2023semantic,
  title={Semantic Anomaly Detection with Large Language Models},
  author={Elhafsi, A. and Sinha, R. and Agia, C. and Schmerling, E. and others},
  journal={Auton. Robots},
  volume={47},
  number={8},
  pages={1035--1055},
  year={2023},
  publisher={Springer}
}

@article{ren2025cot,
  title={Cot-vlm4tar: Chain-of-thought guided vision-language models for traffic anomaly resolution},
  author={Ren, T. and Hu, H. and Zuo, J. and Chen, X. and others},
  journal={arXiv preprint arXiv:2503.01632},
  year={2025}
}

@misc{sinha2024realtimeanomalydetectionreactive,
  title={Real-Time Anomaly Detection and Reactive Planning with Large Language Models},
  author={Sinha, R. and Elhafsi, A. and Agia, C. and Foutter, M. and others},
  year={2024},
  eprint={2407.08735},
  archivePrefix={arXiv},
  primaryClass={cs.RO},
  url={https://arxiv.org/abs/2407.08735}
}

@misc{bai2025qwen25vltechnicalreport,
  title={Qwen2.5-VL Technical Report},
  author={Bai, S. and Chen, K. and Liu, X. and Wang, J. and others},
  year={2025},
  eprint={2502.13923},
  archivePrefix={arXiv},
  primaryClass={cs.CV},
  url={https://arxiv.org/abs/2502.13923}
}

@misc{nvidia2025cosmosreason1physicalcommonsense,
  title={Cosmos-Reason1: From Physical Common Sense To Embodied Reasoning},
  author={{NVIDIA} and Azzolini, A. and Brandon, H. and Chattopadhyay, P. and others},
  year={2025},
  eprint={2503.15558},
  archivePrefix={arXiv},
  url={https://arxiv.org/abs/2503.15558}
}

@inproceedings{dao2024flashattention2,
  title={{FlashAttention-2}: Faster Attention with Better Parallelism and Work Partitioning},
  author={Dao, T.},
  booktitle={Int. Conf. Learn. Represent. (ICLR)},
  year={2024},
  url={https://arxiv.org/abs/2307.08691}
}

@article{arya2024rdd2022,
  title={{RDD2022}: A multi-national image dataset for automatic road damage detection},
  author={Arya, D. and Maeda, H. and Ghosh, S. K. and Toshniwal, D. and others},
  journal={Geosci. Data J.},
  volume={11},
  number={4},
  pages={846--862},
  year={2024},
  publisher={Wiley},
  doi={10.1002/gdj3.260}
}

@inproceedings{cordts2016cityscapes,
  title={The {Cityscapes} Dataset for Semantic Urban Scene Understanding},
  author={Cordts, M. and Omran, M. and Ramos, S. and Rehfeld, T. and others},
  booktitle={IEEE/CVF Conf. Comput. Vis. Pattern Recognit. (CVPR)},
  pages={3213--3223},
  year={2016},
  organization={IEEE}
}

@misc{theorypass_hazard_perception,
  author={{Theory Pass}},
  title={Hazard Perception Practice},
  year={2024},
  howpublished={\url{https://theorypass.co.uk/practice/hazard-perception}},
  note={Accessed: 2026-03-02}
}

@article{yao2020unsupervised,
  title={{DoTA}: Unsupervised Detection of Traffic Anomaly in Driving Videos},
  author={Yao, Y. and Wang, X. and Xu, M. and Pu, Z. and others},
  journal={IEEE Trans. Pattern Anal. Mach. Intell.},
  volume={45},
  number={1},
  pages={444--459},
  year={2023},
  publisher={IEEE}
}

@inproceedings{fang2021dada,
  title={{DADA}: Driver Attention in Driving Accident Scenarios},
  author={Fang, J. and Yan, D. and Qiao, J. and Xue, J. and others},
  booktitle={IEEE/CVF Int. Conf. Comput. Vis. (ICCV)},
  pages={4537--4547},
  year={2021},
  organization={IEEE}
}

@inproceedings{liznerski2021explainable,
  title={Explainable Deep One-Class Classification},
  author={Liznerski, P. and Ruff, L. and Vandermeulen, R. A. and Franks, B. J. and others},
  booktitle={Int. Conf. Learn. Represent. (ICLR)},
  year={2021},
  url={https://openreview.net/forum?id=A5VV3UyIQz}
}
\end{document}